\pdfoutput=1
\documentclass[twocolumn,times,final,authoryear]{elsarticle}

\usepackage{prletters}
\usepackage{framed,multirow}

\usepackage{amssymb}
\usepackage{amsmath}
\usepackage{amsfonts}
\usepackage{bm}
\usepackage{latexsym}

\usepackage{url}
\usepackage{xcolor}
\definecolor{newcolor}{rgb}{.8,.349,.1}

\usepackage[acronym,nomain,nonumberlist]{glossaries}
\usepackage[figuresright]{rotating}
\usepackage[export]{adjustbox}[2011/08/13]
\usepackage[utf8]{inputenc}
\usepackage{subcaption}
\usepackage{placeins}
\usepackage{url}
\usepackage{booktabs}
\usepackage{graphics}
\newacronym{z}{Z}{mean atomic number}
\newacronym{tima}{TIMA}{TESCAN Integrated Mineral Analyzer}
\newacronym{lut}{LUT}{Lappeenranta-Lahti University of Technology}
\newacronym{but}{BUT}{Brno University of Technology}
\newacronym{nn}{NN}{Neural Network}
\newacronym{dbn}{DBN}{Deep Belief Net}
\newacronym{rbm}{RBM}{Restricted Boltzmann Machine}
\newacronym{cnn}{CNN}{Convolutional Neural Network}
\newacronym{vit}{ViT}{Vision Transformer}
\newacronym{mlp}{MLP}{Multi-Layer Perceptron}
\newacronym{slp}{SLP}{Single-Layer Perceptron}
\newacronym{rscnn}{RSCNN}{Relation-Shape Convolutional Neural Network}
\newacronym{gnn}{GNN}{Graph Neural Network}
\newacronym{rgbd}{RGB-D}{RGB + depth}
\newacronym{svm}{SVM}{Support Vector Machine}
\newacronym{mri}{MRI}{Magnetic Resonance Imaging}
\newacronym{fnn}{FNN}{Feedforward Neural Network}
\newacronym{sfcnn}{SFCNN}{Spherical Fractal Convolutional Neural Network}
\newacronym{kNN}{kNN}{k-Nearest Neighbours}
\newacronym{gat}{GAT}{Graph Attention Network}
\newacronym{gcn}{GCN}{Graph Convolutional Network}
\newacronym{gin}{GIN}{Graph Isomorphism Network}
\newacronym{dt}{DT}{Delaunay triangulation}

\newacronym{sem}{SEM}{Scanning Electron Microscope}
\newacronym{eds}{EDS}{Energy-Dispersive X-Ray Spectroscopy}
\newacronym{bse}{BSE}{Backscattered Electrons}

\newacronym{roc}{ROC}{Receiver Operating Characteristic}
\newacronym{auroc}{AUROC}{Area Under the Receiver Operating Characteristic}
\newacronym{iou}{IoU}{Intersection over Union}
\newacronym{tp}{TP}{True Positive}
\newacronym{tn}{TN}{True Negative}
\newacronym{fp}{FP}{False Positive}
\newacronym{fn}{FN}{False Negative}
\newacronym{gdls}{GDLS}{Graph-based Deep Learning Segmentation}
\newacronym{yolo}{YOLO}{You Only Look Once}
\newacronym{ct}{CT}{Computed Tomography}

\newacronym{mla}{MLA}{Mineral Liberation Analyzer}
\newacronym{qemsem}{QEM*SEM}{Quantitative Evaluation of Minerals by Scanning Electron Microscopy}
\graphicspath{{figures/},{resultfigures/}}

\journal{Pattern Recognition Letters}

\begin{document}

\setcounter{page}{1}

\begin{frontmatter}

\title{Mineral segmentation using electron microscope images and spectral sampling through multimodal graph neural networks}

\author[lut,but]{Samuel Repka\corref{cor1}} 
\ead{samuel.repka@lut.fi}
\author[lut,but]{Bořek Reich} 
\ead{borek.reich@lut.fi}
\author[lut]{Fedor Zolotarev} 
\ead{fedor.zolotarev@lut.fi}
\author[lut]{Tuomas Eerola} 
\ead{tuomas.eerola@lut.fi}
\author[lut,but]{Pavel Zemčík} 
\ead{pavel.zemcik@lut.fi}

\affiliation[lut]{organization={Lappeenranta-Lahti University of Technology},
            addressline={Yliopistonkatu 34}, 
            city={Lappeenranta},
            postcode={53850}, 
            country={Finland}}

\affiliation[but]{organization={Brno University of Technology - Faculty of Information Technology},
            addressline={Božetěchova 1/2}, 
            city={Brno},
            postcode={61266}, 
            country={Czechia}}

\cortext[cor1]{Corresponding author
  }

\begin{abstract}
We propose a novel Graph Neural Network-based method for segmentation based on data fusion of multimodal Scanning Electron Microscope (SEM) images. In most cases, Backscattered Electron (BSE) images obtained using SEM do not contain sufficient information for mineral segmentation. Therefore, imaging is often complemented with point-wise Energy-Dispersive X-ray Spectroscopy (EDS) spectral measurements that provide highly accurate information about the chemical composition but that are time-consuming to acquire. This motivates the use of sparse spectral data in conjunction with BSE images for mineral segmentation. The unstructured nature of the spectral data makes most traditional image fusion techniques unsuitable for BSE-EDS fusion. We propose using graph neural networks to fuse the two modalities and segment the mineral phases simultaneously. Our results demonstrate that providing EDS data for as few as 1\% of BSE pixels produces accurate segmentation, enabling rapid analysis of mineral samples. The proposed data fusion pipeline is versatile and can be adapted to other domains that involve image data and point-wise measurements. 
\end{abstract}


\end{frontmatter}


\section{Introduction}
\label{sec:intro}
Multimodal image segmentation has been widely studied~\citep{Zhang_multimodalSegmentationSurvey_2021} due to its diverse applications, ranging from medical image analysis~\citep{Guo_MedicalSegmentation_2019} to self-driving cars~\citep{Feng_SelfDrivingFusion_2021} and various industrial processes~\citep{Itou_MultimodalIndustrial_2023}. Leveraging additional modalities to complement image data can significantly increase segmentation accuracy. When the modalities share a similar structure (e.g., RGB images and depth maps both having a regular grid-like structure), various common fusion methods can be employed, which has led to a number of well-established multimodal segmentation techniques. However, this is not the case when the modalities are structurally different, such as data sampled in a non-grid-like fashion. A good example of this is image/spectral data, such as \gls{sem} imaging accompanied by a \gls{eds}, which produces images and point-wise spectral measurements. While the benefits of accurate spectral measurements are evident, fusing them with image data for segmentation is not straightforward.

Knowledge of the chemical composition of a specimen on a microscale is essential for various fields, such as mineralogy or geology. \gls{sem} with \gls{bse} and \gls{eds} detectors is a commonly used process to acquire data describing the chemical composition \citep{Girao_2017_EDS}. A specimen with a known chemical composition and reasonable resolution can then be segmented into different phases or grains of material with the same chemical composition.
Accurate mineral segmentation is crucial for quantifying mineral abundance, measuring size and shape, and analysing spatial relationships. This data is essential for further analysis and interpretation.
The measurements are applied abundantly in mineralogy, geology, and geometallurgy in tasks such as plant optimisation, research, ore characterisation, and/or process circuit surveys~\citep{Gottlieb_applications_2000}. Other applications include uses in environmental sciences, petrology, and recycling industry~\citep{Sandmann_applications_2015}.
One way to perform \gls{sem} image segmentation is to use only \gls{bse} data. The \gls{bse} sensor produces a greyscale image, with grey levels proportional to \gls{z}. The main benefit of this approach is the speed of data acquisition, as high-quality \gls{bse} data can be acquired efficiently. However, various samples may have different chemical compositions but similar \gls{z}, which limits the segmentation ability of methods using only \gls{bse} data. An alternative segmentation approach is to utilise the \gls{eds} data, which contain more versatile information on the chemical composition of the sample. 
The \gls{eds} data can be used to segment the sample into different phases accurately. 
However, the acquisition of \gls{eds} data is much more time-consuming (in standard cases 1 \textmu s per \gls{bse} pixel and 1.25 ms per \gls{eds} spectrum), making the analysis more expensive.
This can be alleviated by acquiring only sparse \gls{eds} data. The sparsity of data means that grain contours cannot be determined accurately. A possible solution is to sample more densely in the contour areas (creating the "unstructured data" problem) or to use dense \gls{bse} and sparse \gls{eds} data simultaneously via data fusion. 



\begin{figure}[!htp]
    \centering
    \includegraphics[width=1\linewidth]{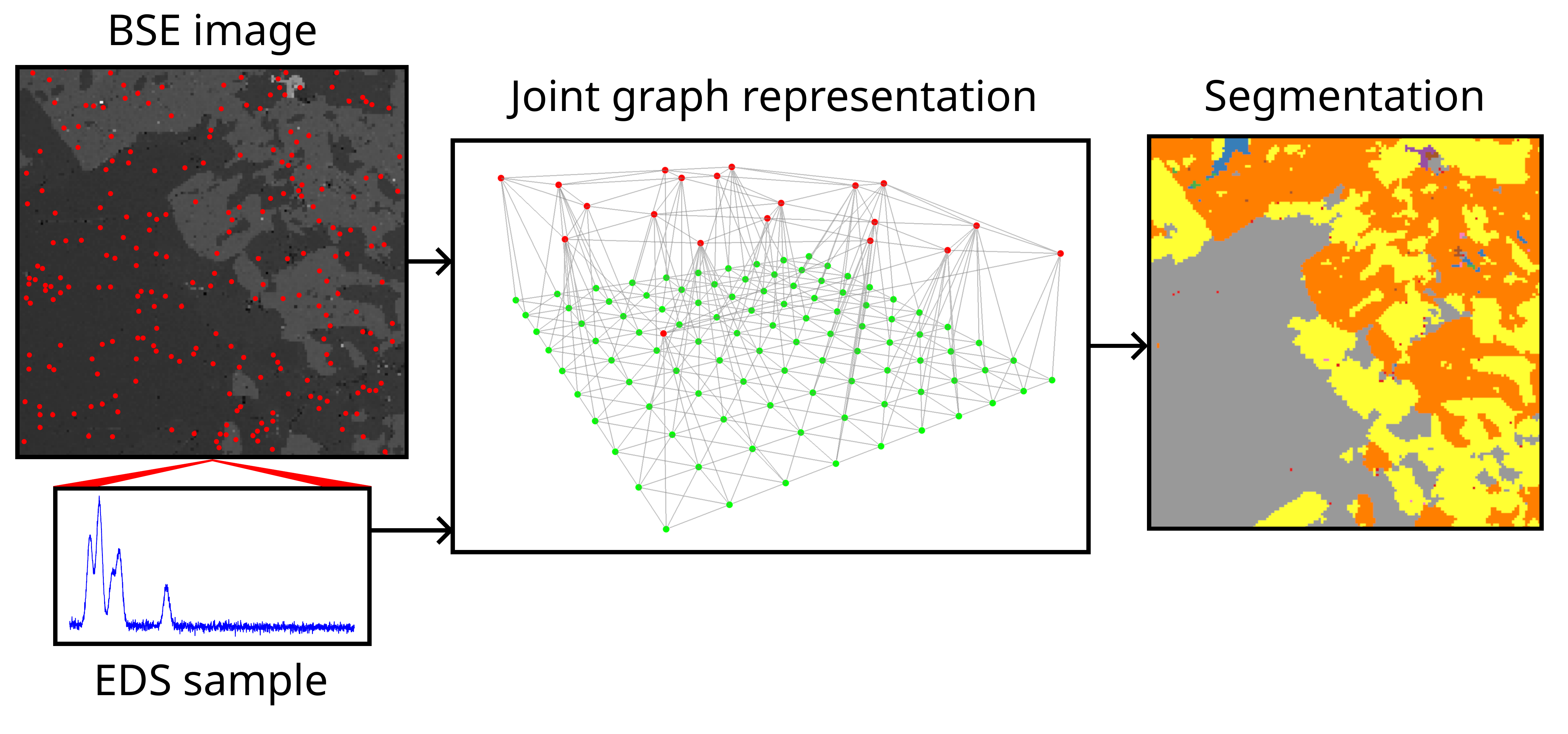}
    \caption{Illustration of the proposed method.}
    \label{fig:scheme}
\end{figure}

In this paper, we approach the problem by introducing a novel method for image segmentation utilising a graph representation and deep learning, illustrated in Fig.~\ref{fig:scheme}. 
Graphs provide a flexible approach for data fusion capable of processing various types of modalities (e.g., images, point clouds, point-wise measurements) without additional preprocessing. The core of the proposed \gls{sem} image segmentation method lies in a graph construction that represents the raw \gls{bse} data and sparse \gls{eds} data and their subsequent processing using a \gls{gat}~\citep{Velickovic_GAT_2018}. 
The proposed fusion pipeline is versatile and can be adapted to other applications involving image and unstructured data, such as point-wise measurements.

In the experimental part of the work, we compare the proposed method to a baseline UNet-based segmentation method applied to \gls{bse} images, demonstrating the benefits of \gls{bse}-\gls{eds} fusion. We further compare the method with the current state-of-the-art approach~\citep{Juranek_rgbEDS_2022}. We show that the proposed method performs better in almost all test cases.

The main contributions of the paper are as follows:
(a) a novel method to construct a combined graph representation of image data and spatially aligned non-grid-like data, 
(b) a segmentation method using Graph Neural Networks, which does not require user-set parameters, and
(c) the application of the proposed solution on mineral segmentation from multimodal \gls{sem} data, demonstrating higher accuracy than
competing methods.

\section{Related work}
\label{sec:related}

\subsection{Multimodal segmentation}
\label{sec:segm}

Data fusion is widely used in image segmentation, with neural networks being the dominant approach. 
Generally, fusion methods can be divided into (1) early fusion, (2) late fusion, and (3) hybrid fusion methods~\citep{Zhang_multimodalSegmentationSurvey_2021}.


In early fusion methods, the data is fused in raw form or at the feature level \citep{Zhang_multimodalSegmentationSurvey_2021}. Raw-form fusion was utilised, for example, by \citet{Couprie_indoorSegmentation_2013}, where the first attempt at deep multimodal fusion was proposed by jointly processing RGB and depth images with \glspl{cnn}. The possibilities of fusion of RGB and sparse depth data using direct channel concatenation were explored by \citet{Jaritz_Segmentation_2018} to perform semantic segmentation and depth completion.
Feature-level fusion is often done by encoding both modalities separately in respective encoders, with cross-modal interactions in the encoding stage. This approach is used in FuseNet \citep{Hazirbas_FuseNet_2017} with \gls{rgbd} data and the RGB and thermal data fusion method by \citet{Sun_RGBthermal_2019}.


The paradigm of late fusion is to integrate the feature maps of the modalities at the decision level. Thus, data must first be processed separately, omitting any information sharing between feature extractors.
A representative sample of this category is the work of \citet{Gupta_lateFusion_2014}. Two separate neural networks are used to extract features from RGB and depth data, which are combined by \gls{svm}. PIF-Net by \citet{Guo_PIFnet_2023} is another example of such a network, adapted for the fusion of images and point clouds, capable of semantic segmentation for both input modalities. It uses two encoders and two decoders, but only encoders share information.

Hybrid fusion methods are meant to alleviate the shortcomings of early and late fusion methods. Usually, in the case where neural networks are used, skip connections are employed to bridge the encoders and a decoder to improve the segmentation performance. An example of this approach is a method proposed by \citet{Lee_hybrid_2017}, or work by \citet{Fang_hybridMRI_2022}, where hybrid data fusion is used to segment brain tumours from \gls{mri} modalities.


\subsection{Mineral segmentation for SEM images}
\label{sec:mineral}
Various methods of automated mineralogy have been developed. Traditional methods include watershed-based methods~\citep{Motl_Watershed_2015}, \gls{mla}~\citep{Fandrich_MLA_2007}, and \gls{qemsem}~\citep{Miller_QEMSEM_1983}. Currently, many commercial solutions are available, such as \acrfull{tima}, Zeiss Mineralogic, Maps Min by ThermoFisher or Bruker's AMICS. However, the workings of algorithms are frequently proprietary and not publicly known.

An alternative approach to the use of both \gls{bse} and \gls{eds} data based on spectral pansharpening was proposed in \citep{Duma_Fusion_2022,Sihvonen_Fusion_2024}. The approach was adapted from satellite data processing and it attempts to increase \gls{eds} resolution with acquired image data. The method is made to improve the \gls{eds} resolution, which is inherently lower than the resolution of an image, but it does not consider the unstructuredness of \gls{eds} data.


\citet{Juranek_rgbEDS_2022} proposed another method for mineral segmentation from \gls{sem} images. The method utilises a joint graph representation created from \gls{bse} and reduced \gls{eds} spectra.  
First, the input EDS spectra are transformed by a \gls{cnn} to obtain a compact descriptor for each spectrum. With this reduced representation, a graph is built by Voronoi analysis, with vertices being assigned a spectrum descriptor, \gls{bse} value and its location. The edges are assigned two values $\delta^b$ and $\delta^e$, capturing differences between \gls{bse} values and \gls{eds} descriptors, respectively. Edges with $\delta$ values greater than user-set thresholds are removed. The resulting graph components serve as a starting point for Markov Random Field segmentation, which produces final segments.
It is worth noting that the resulting segments do not have any phase assigned; they just constitute parts of the sample with sufficiently similar chemical composition. The algorithm has several drawbacks, for example, segmentation dependence on two parameters, which need to be set by an expert, or high computational complexity.

Maps Min software~\citep{Han_ThermoFisher_2022} utilises \gls{eds} spectra to produce segmentation, with a focus on potential mixed-material spectra. The proprietary "Mixel algorithm" produces pixel classification for up to three mineral phases.
AMICS by Bruker~\citep{brukerSoftwareAMICS} can operate in multiple modes, segmentation and mapping mode. In segmentation mode, the solution attempts to minimise the number of EDS points necessary to characterise a sample. A mapping mode collects \gls{eds} data in an equidistant grid and creates grains by combining similar measurements.

\subsection{Graph neural networks}
\label{sec:gnn}

Graphs are versatile data structures, suitable for processing structured as well as unstructured data, such as point-wise \gls{eds} measurements. To efficiently exploit the graph representation in computation, various methods have been created to process such data. A modern approach is to use specialised neural networks.


A \gls{gnn} applies a differentiable model 
to all components of the graph (nodes, edges, and global embeddings), transforming them into a new graph. Data integration between neighbouring nodes (or edges, but not node to edge or edge to node) is facilitated by \emph{pooling} \citep{Sanchez_GNN_2021}. Similarly to the convolution in \gls{cnn}, for items to be pooled, data are first gathered and then aggregated by a function, the simplest of which are sum or mean. 
In order to combine information from different parts of the graph (e.g., edges to nodes), a \emph{message passing layer} is used. Due to the high flexibility and applicability of graphs, it is not a surprise that the architectural landscape of \glspl{gnn} is quite diverse. Variants include \acrlong{gcn} \citep{Kipf_GCN_2017}, \gls{gat} \citep{Velickovic_GAT_2018} and \acrlong{gin} \citep{Xu_GIN_2019}.

\glspl{gnn} have been successfully used in various tasks including protein folding~\citep{Gligorijevic_GNNprotein_2021}, physics simulation~\citep{Gonzales_GNNphysics_2020}, various computer vision tasks~\citep{Chen_GNNCV_2024}, and more~\citep{Sanchez_GNN_2021}.

\section{Proposed method}
\label{sec:proposed}

The proposed method for \gls{bse}-\gls{eds} data fusion and segmentation is based on a graph representation of both modalities, with edges constructed to leverage spatial relationships between data nodes. The graph is then processed using a \acrfull{gat}, to obtain a fused representation of the modalities. The final output of the method is, in this case, the image part of the graph containing the resulting segmentation.

\subsection{Pipeline}
\begin{figure}[!htp]
    \centering
    \includegraphics[width=1\linewidth]{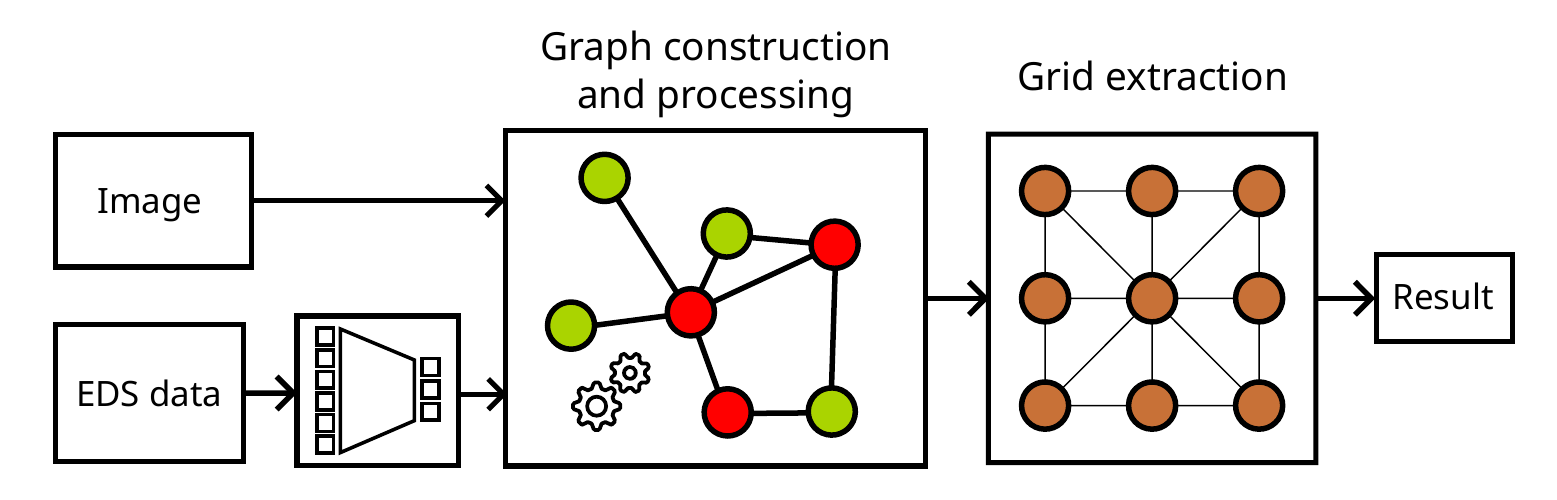}
    \caption{Proposed pipeline. Firstly, the dimensionality of \gls{eds} modality is reduced and then, together with image data, are put into a graph. The graph is then processed with a \gls{gnn}. The part of the graph representing the image is extracted, which represents the segmented image.}
    \label{fig:framework}
\end{figure}

The overview of the solution is depicted in Fig.~\ref{fig:framework}. It revolves around a joint representation of both modalities as a graph and data fusion by a graph processing technique. 
First, the dimensionality of the raw \gls{eds} data is reduced to obtain compact-size embeddings. For this, we use \gls{cnn}-based embedding method by \citet{Juranek_rgbEDS_2022}.
The next step is \textit{graph construction}, where a graph is constructed from both modalities, serving as a joint intermediate representation. 
In the \textit{graph processing} step, the data is fused to produce either a segmentation result or fused data representation. For this, we propose the utilisation of \acrfull{gat}.
Finally, the graph part representing the image is extracted from the graph, producing the final segmentation mask.


\subsection{Graph construction}
Since the final goal is image segmentation, the output of the fusion framework needs to be grid-like and has the same structure as the input image. Thus, a method for the transformation of the \gls{eds} data into a structured format needs to be devised.

The proposed method assumes that a spatial relationship exists between modalities and that both modalities are aligned. For example, in the case of image and point-wise measurements, it is assumed the measurements correspond to the known spatial locations (pixels) in the image. This alignment allows for a meaningful fusion of the two modalities, as spatial relationships can be exploited to improve the segmentation results.

To stay as general as possible, a decision was made to represent both modalities in a joint form using a graph. This allows for the exact spatial representation and also allows for "weighing" of sample points with respect to each other. This can be achieved by assigning graph edges connecting points a number expressing how likely are those measurements correlated. 

\subsubsection*{Edge construction}
Edges of the graph should be constructed in a way, which allows for the optimal leveraging of spatial relationships between data nodes. Due to the properties of modalities, edges can be split into three subgroups: (1) edges between the nodes (pixels) of the image, (2) edges between nodes of the \gls{eds} modality, and (3) edges between nodes of different modalities.


The straightforward way to present an image as a graph is to consider pixels as nodes and to create edges between adjacent pixels. As the image is always structured, the direct solution is to create edges connecting neighbouring pixels in 4 or 8-neighbourhood. Here, a choice was made to connect the adjacent pixels using an 8-neighbourhood.

\begin{figure}[ht!]
    \centering
    \begin{subfigure}{0.48\columnwidth}
        \includegraphics[width=\linewidth]{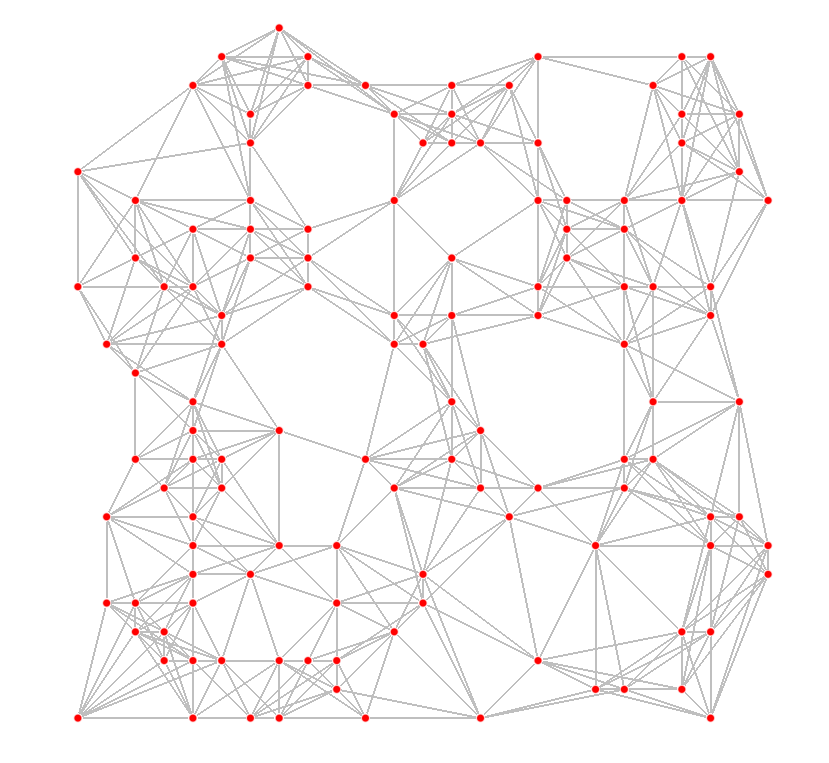}
        \caption{Graph constructed using 8-nearest neighbours.}
    \end{subfigure}
    \begin{subfigure}{0.48\columnwidth}
        \includegraphics[width=\linewidth]{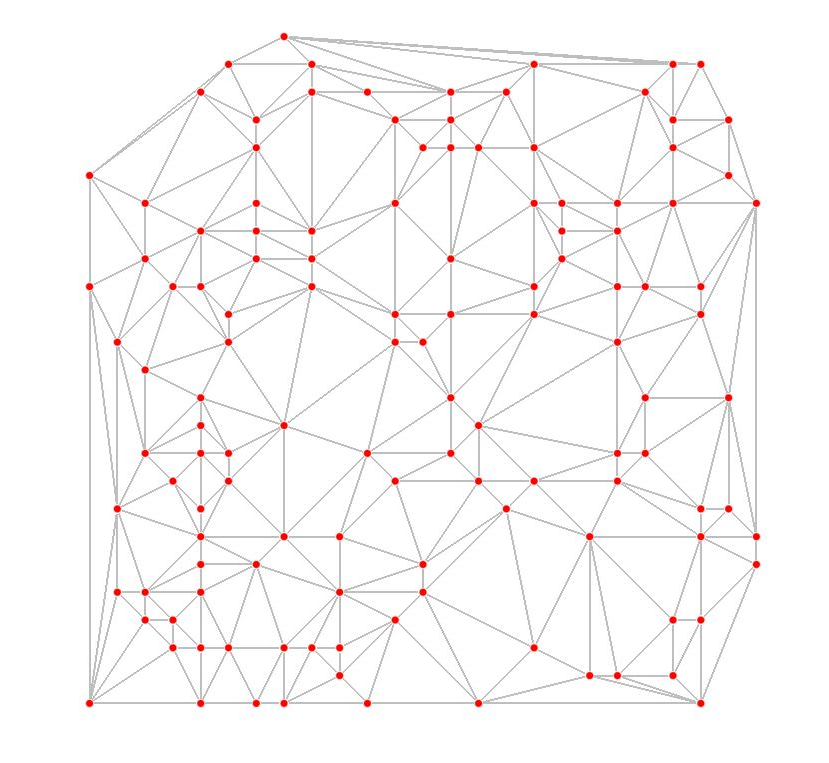}
        \caption{Graph constructed with \gls{dt}.}
    \end{subfigure}
    \caption{Different types of graph constructions from point-wise measurements.}
    \label{fig:non_structured_edges}
\end{figure}

The \gls{eds} modality is unstructured, which complicates the creation process of the second edge type, as it is not unambiguous about which nodes should share an edge. One option is to use a \gls{kNN} graph \citep{Dong_KNNG_2011} where, for each node, edges are formed to $k$ most similar nodes based on, e.g., Euclidean distance. This, however, can create edges that are oriented in a narrow angle in a direction, if the closest nodes are in a cluster close to each other. It can also increase the graph diameter, which hinders the propagation of information. Moreover, in extreme cases, it can create isolated clusters of nodes. To obtain a cleaner graph representation, the proposed method uses \gls{dt} \citep{deBerg_delaunay_2008} to construct edges. \gls{dt} is a method of creating a triangulation of a set of points where no point is inside the circumcircle of any triangle. It maximises the smallest angle in any of the triangles, which leads to a more uniform distribution of edges.
%
%
Examples of triangulation using both \gls{kNN} graph and \gls{dt} can be seen in Fig.~\ref{fig:non_structured_edges}. As it can be seen, \gls{dt} produces a simpler graph with edges spread all around most of the points, which allows for simpler usage of information from diverse neighbourhoods. The two approaches were also experimentally evaluated on the proposed pipeline. \gls{dt} outperformed the \gls{kNN} method in all measured metrics. Details can be seen in Sec. \ref{sec:results}.

\begin{figure}[!htp]
    \centering
    \includegraphics[width=0.9\linewidth]{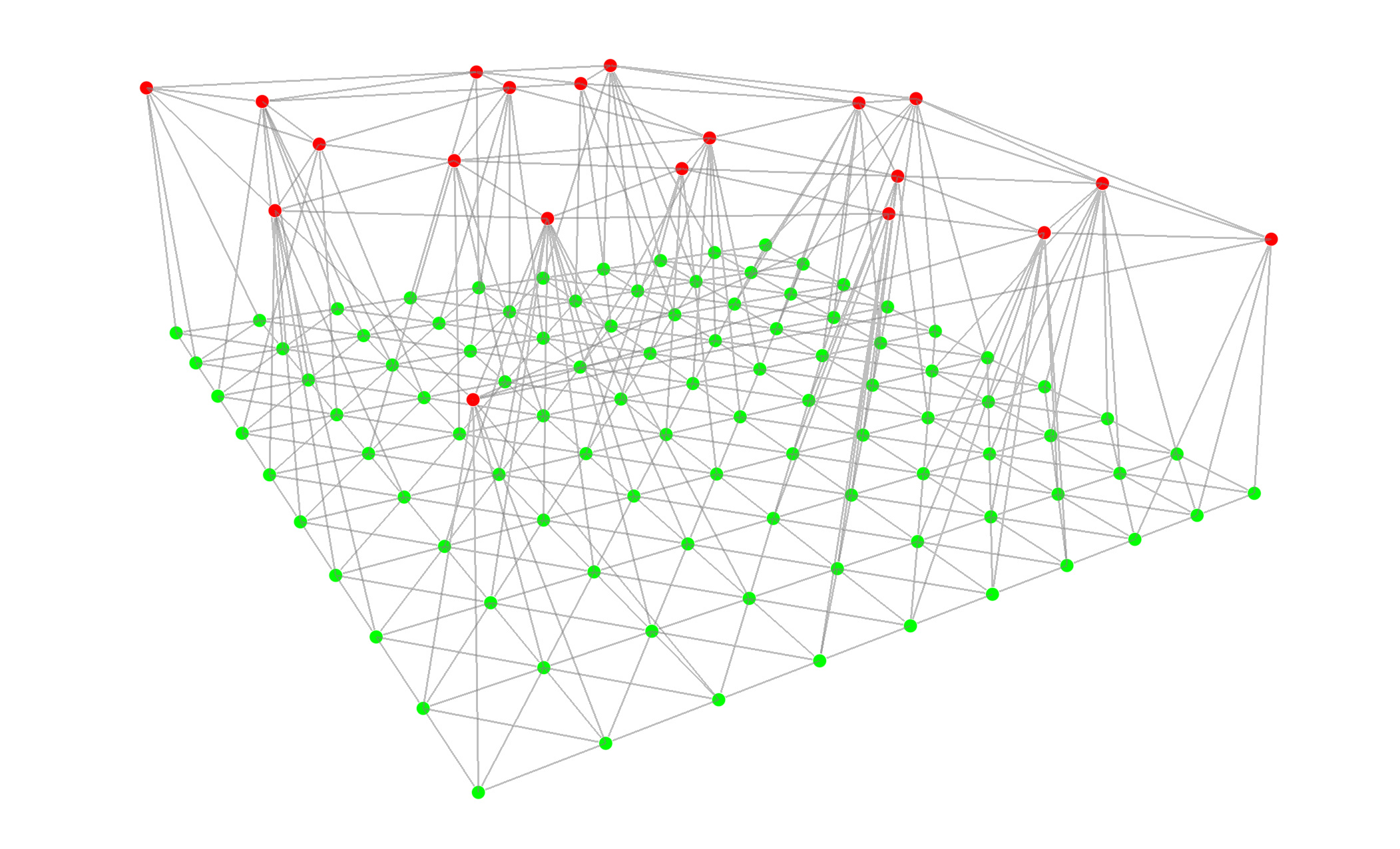}
    \caption{Illustration of a fully constructed graph with edges between both modalities. Green points represent the image and red points the \gls{eds} modality.}
    \label{fig:complete_graph}
\end{figure}

The third type of edge connects nodes of different modalities, which are spatially close to each other. It is assumed that the modalities are aligned (i.e. the $xy$-coordinates of one modality correspond to the same world position as the same coordinates in the second modality). To keep things concise, let $M1$ denote the graph representing the first (image) modality and $M2$ denote the graph representing the second (unstructured) modality. The edge creation process begins by placing both graphs into a one-higher dimensional space (enhancing $xy$-coordinates of measurement locations with a third coordinate). In the case of \gls{bse}-\gls{eds} data, the two modalities are put "above" each other. Here, a choice was made for the additional spatial dimension to be 0 for $M1$ and 1 for $M2$. Next, an edge is created for every node of $M1$ with the closest node of $M2$. To ensure that no point is left without an edge, the same is done vice versa, and duplicates are later deleted. Fig.~\ref{fig:complete_graph} shows the resulting graph.

\subsubsection*{Edge  attributes}
All edges are added an attribute (or weight), in the form of its Euclidean distance between connected nodes.  Note that the choice of values of additional dimension when constructing the third edge type heavily affects the attributes in relevant edges. As it is assumed that a minimum distance of 2 different \gls{bse} nodes is 1 (because of the discrete image grid), the values ensure that this also holds in the case of the third edge type. Moreover, it ensures that self-loops are properly distinguished from the rest of the edges.

\subsubsection*{Node attributes}
The possible non-commensurability of modalities means that there is no straightforward approach for the representation of both modalities in a single graph. Two main possibilities were considered: (1) a heterogeneous graph consisting of two types of nodes, and (2) zero padding, where only one type of node exists and features from both modalities are concatenated, filling zeroes to places where data from either modality is not available.
The second option was chosen because of its relative simplicity and the fact that it allows for the usage of the existing algorithms developed for the homogeneous graphs. Applied to \gls{bse} and \gls{eds} data, the node attributes contain a 65-tuple, where the first value represents the \gls{bse} measurement and the remaining 64 values represent reduced \gls{eds} spectra. In the case of nodes representing \gls{bse} pixels, the 64 values would be set to 0, and similarly for \gls{eds} nodes, the first value would be 0.

\subsection{Graph processing}
A \gls{gnn} is used to process the graph. The proposed method utilises \gls{gat}~\citep{Velickovic_GAT_2018}, because of its ability to capture spatial relationships between nodes, as well as the expectation that the attention mechanism would be able to distinguish between important and less important graph edges. An implementation from the PyTorch Geometric library \citep{Fey_PyG_2019} was used, which also incorporated edge attributes into the graph processing. The network produces a new set of node features $\mathbf{x}'_i$ for $i$-th node from the original set $\mathbf{x}_i$ as follows:
\begin{equation}
    \mathbf{x}'_i = \sigma \left( \sum_{j \in \mathcal{N}_i} \alpha_{ij} \mathbf{W}_t \mathbf{x}_j \right),
\end{equation}
where $\sigma$ represents a nonlinear activation function, $a_{ij}$ attention coefficients, and $\mathbf{W}_t$ a weight matrix. The attention coefficients are computed as
\begin{equation}
\alpha_{ij} =
\frac{\exp \left( \sigma_{lrl} \left( \bm{a}_s^{T} \mathbf{W}_t \mathbf{x}_i + 
\bm{a}_t^{T} \mathbf{W}_t \mathbf{x}_j + \bm{a}_e^{T} \mathbf{W}_e \mathbf{e}_{ij} \right) \right)}
{\sum\limits_{k \in \mathcal{N}_i} \exp \left(  \sigma_{lrl} \left( 
\bm{a}_s^{T} \mathbf{W}_t \mathbf{x}_i + \bm{a}_t^{T} \mathbf{W}_t \mathbf{x}_k + 
\bm{a}_e^{T} \mathbf{W}_e \mathbf{e}_{ik} \right) \right)}
\end{equation}
where  $\sigma_{lrl}$ is a leaky ReLu activation function, $\mathcal{N}_i$ is a neighbourhood of node $i$ in the graph (including the node itself), $\bm{a}$ are weight vectors, $\mathbf{W}_e$ a weight matrix concerning edge attributes, and $\mathbf{e}_{ij}$ attribute of an edge connecting nodes $i$ and $j$.

Even though in this work only \gls{gat} was used, the method is not restricted to this architecture; any graph processing technique can be used. The output of the processing can be a joint representation of modalities, an augmented version of a single modality, or even direct segmentation. 

The part of the graph representing an image is to be extracted after the processing, obtaining a grid-like representation of data or a result. For example in Fig.~\ref{fig:complete_graph}, the green part of the graph would be extracted. In this case, the \gls{gat} was expected to produce direct segmentation to phases, meaning the final output graph has the same structure, but node attributes are transformed into an array of 50 numbers, each number corresponding to the likelihood of the node corresponding to a given class.

The network consists of 3 layers, each with 56 hidden channels and 4 attention heads. Moreover, while processing the graph, self-loop edges were added with weight 0 to allow the node to pass messages to itself with the highest priority. The last layer output was a 50-channel vector for each pixel, each presenting a single mineral class from the dataset.

\section{Experiments}
\label{sec:exp}
The experiments aimed to evaluate the effectiveness of a proposed method and data fusion. Using a dataset of quartz samples, the study compared the proposed approach to a UNet baseline and the \gls{gdls} method. The goal was to assess how well the model could learn from limited \gls{eds} data and improve segmentation accuracy.

\subsection{Data}
\label{sec:data}

\begin{figure}[ht]
    \centering
    \includegraphics[width=1\linewidth]{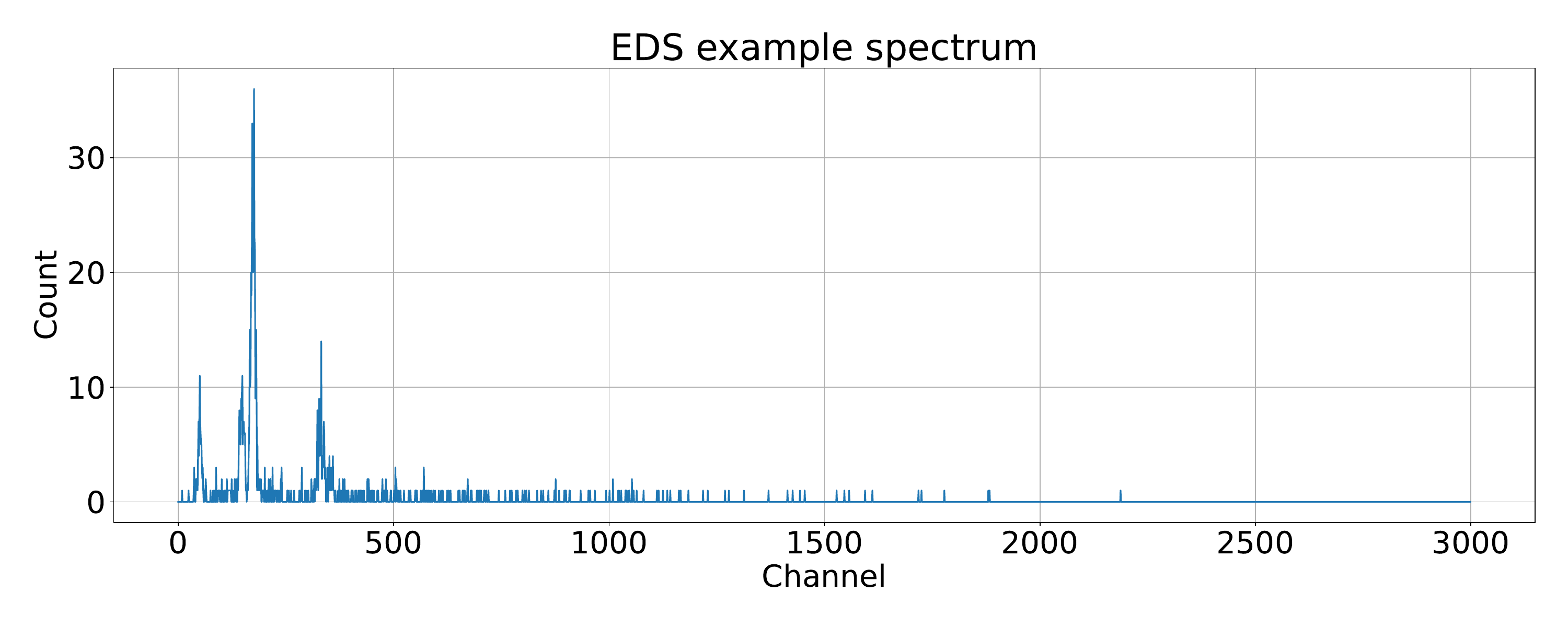}
    \caption{Example of a single pixel EDS spectrum.}
    \label{fig:eds_example}
\end{figure}

The experiments were carried out using the data from~\citep{Breitner_Rock_2022}.
The dataset contains a scan of a quartz sample from the Pitinga deposit. 
The scan was performed with \gls{sem} Tescan TIMA. It contains various modalities, of which two were used: \gls{bse} images and \gls{eds} spectral measurements. The dataset contains 1596 measurements of the various non-intersecting locations on the sample. Each measurement has a $150\times150$ pixel \gls{bse} image (one channel), with the corresponding \gls{eds} spectrum for each pixel. The spectrum represents photon energies from 0 to 30 000 eV in 10 eV wide channels, so a single spectrum is an array of 3000 channels per pixel. An example of such a spectrum can be seen in Fig.~\ref{fig:eds_example}. 

\begin{figure}[ht]
    \centering
    \begin{subfigure}{0.43\columnwidth}
        \includegraphics[width=\linewidth]{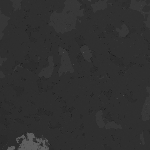}
        \caption{\gls{bse} image from the dataset.}
    \end{subfigure}
    \begin{subfigure}{0.43\columnwidth}
        \includegraphics[width=\linewidth]{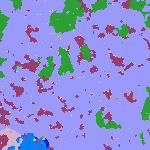}
        \caption{Segmented image.}
    \end{subfigure}
    
    \caption{Example \gls{bse} image with its segmentation to mineral phases.}
    \label{fig:bse_phases}
\end{figure}

Because the number of channels is so high, the method employs a dimensionality reduction to 64 values. The dimensionality reduction was performed by a \gls{cnn} used and created by \cite{Juranek_rgbEDS_2022}. Furthermore, the samples underwent liberation analysis in Tescan TIMA software, producing a mineral phase map of the sample. The map contains the segmentation of the \gls{bse} image, segmenting the minerals in the sample, which was used as a ground truth for the segmentation task. The maps were prepared by a mineralogy expert with a hand-crafted classification schema to ensure correctness and validity.
In the whole dataset, there are 50 distinct classes of minerals. An illustration of \gls{bse} image with the corresponding ground truth segmentation mask can be seen in Fig.~\ref{fig:bse_phases}. During training, the \gls{bse} modality was used as-is, while the \gls{eds} modality was decimated to contain anywhere between 0\% and 70\% of the original data. The dataset was split into training, validation, and testing sets, with 80\% of the data used for training, 10\% for validation, and 10\% for testing. 

One thing to note is that in a few cases, the measurements may be invalid. For example, when an edge of the sample is scanned, the spectral measurements from "beyond the edge" are not valid and may not contain any data at all. These areas were excluded from the evaluation. 

\subsection{UNet Baseline}
\begin{figure}[ht!]
    \centering
    \begin{subfigure}{0.326\columnwidth}
        \includegraphics[width=\linewidth]{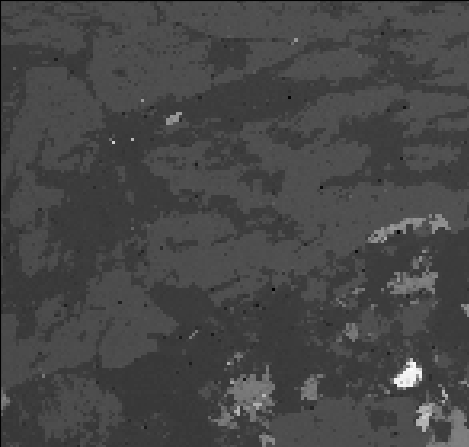}
        \caption{Input BSE image.}
    \end{subfigure}
    \hfill
    \begin{subfigure}{0.326\columnwidth}
        \includegraphics[width=\linewidth]{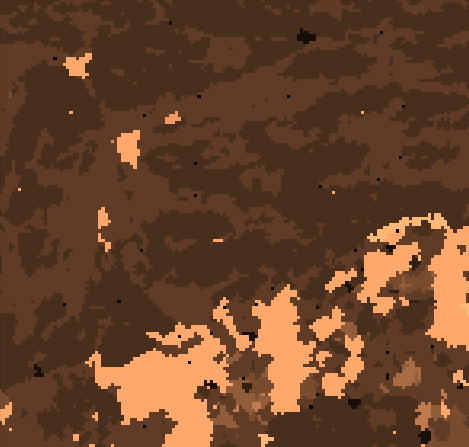}
        \caption{Ground truth.}
    \end{subfigure}
    \hfill
    \begin{subfigure}{0.326\columnwidth}
        \includegraphics[width=\linewidth]{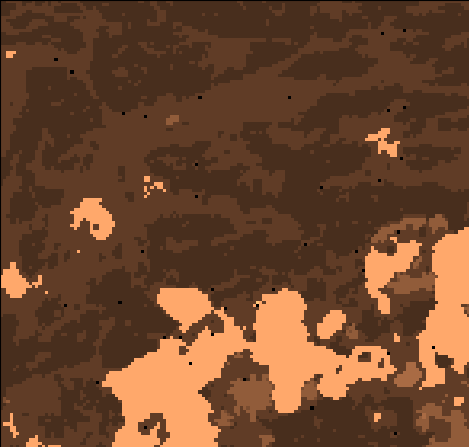}
        \caption{Prediction.}
    \end{subfigure}
    \caption{Example data and output of the UNet baseline.}
    \label{fig:unet_example}
\end{figure}

The UNet \citep{Ronneberger_U-Net_2015} trained exclusively on \gls{bse} data was used as a baseline to demonstrate the segmentation accuracy without \gls{eds} data and data fusion. Example outputs are shown in Fig.~\ref{fig:unet_example}. As can be seen, the UNet fails to capture some of the finer details and completely misclassifies certain regions or minerals. 

\begin{figure}[ht!]
    \centering
    \begin{subfigure}{0.326\columnwidth}
        \includegraphics[width=\linewidth]{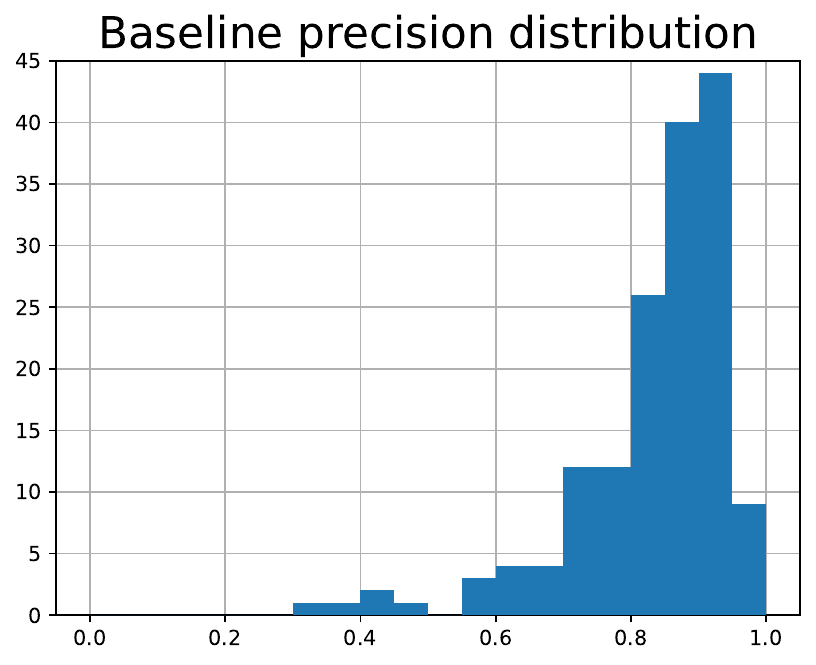}
    \end{subfigure}
    \hfill
    \begin{subfigure}{0.326\columnwidth}
        \includegraphics[width=\linewidth]{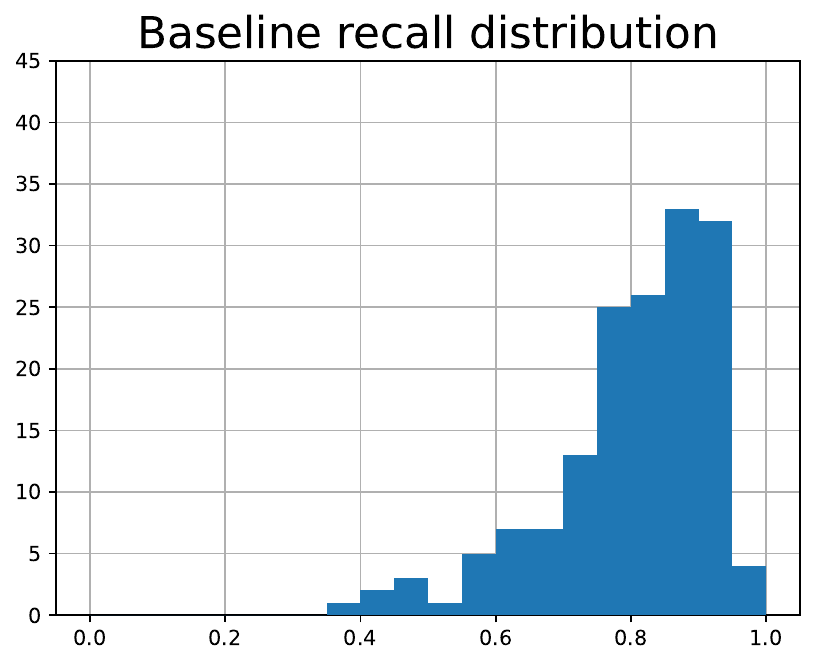}
    \end{subfigure}
    \hfill
    \begin{subfigure}{0.326\columnwidth}
        \includegraphics[width=\linewidth]{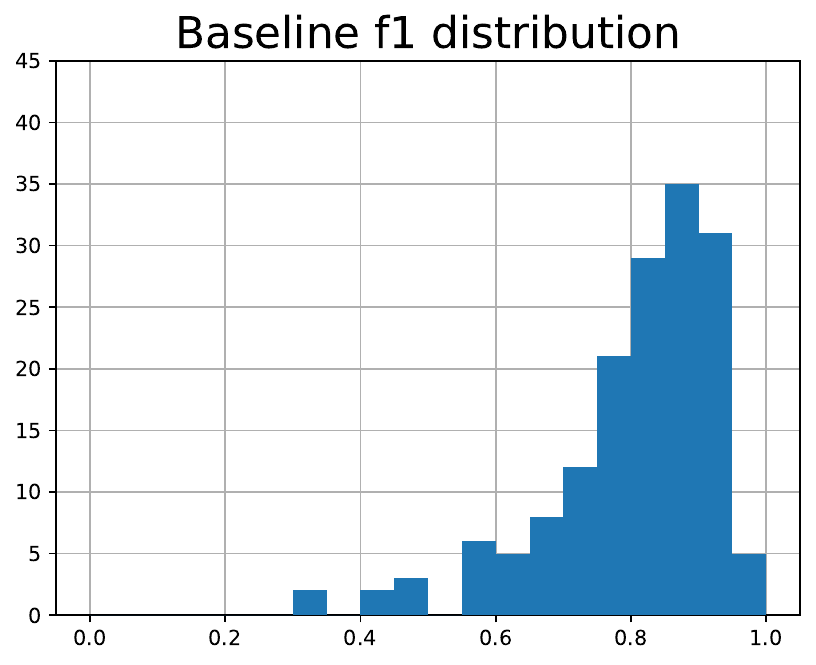}
    \end{subfigure}
    \caption{Histograms of UNet baseline metrics over the test set.}
    \label{fig:baseline_metrics}
\end{figure}

The means of precision, recall, and F1 score for UNet were 0.835, 0.804, and 0.806, respectively. These results are used in further evaluation as a comparison to a naive solution. Distributions of metrics over the test set can be seen in Fig.~\ref{fig:baseline_metrics}.

\subsection{Hyperparameters}
\begin{figure}[ht!]
    \centering
    \includegraphics[width=\columnwidth]{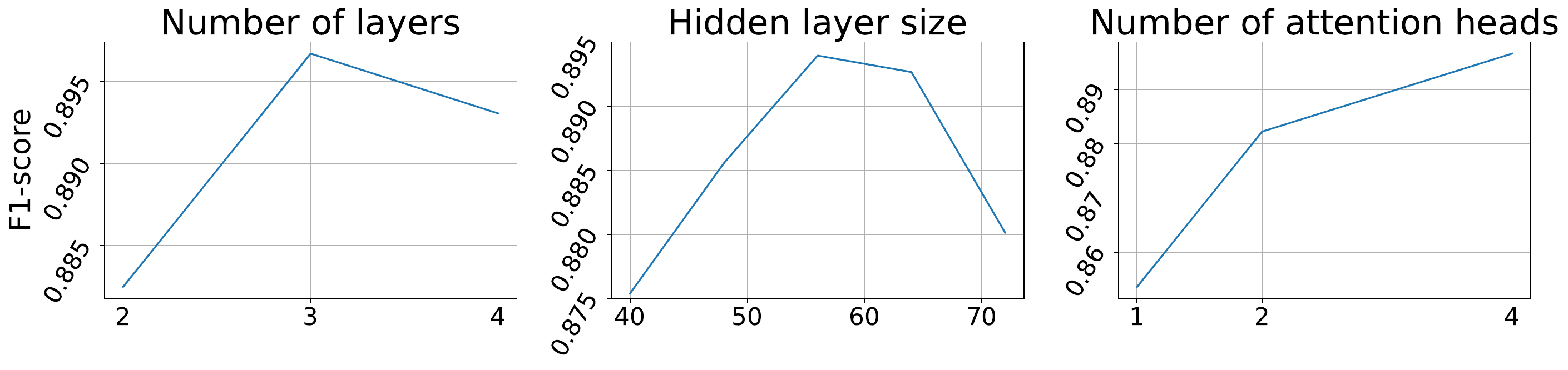}
    \caption{Evaluation of hyperparameters.}
    \label{fig:hyperparameters}
\end{figure}

An extensive hyperparameter optimisation was performed for three main hyperparameters: number of layers, hidden layer size, and number of attention heads. Each model was trained with a batch size of 16 and a learning rate of 0.01 over 120 epochs. 
Fig.~\ref{fig:hyperparameters} shows the performance metric for various models in the test set. Based on the results, the following hyperparameters were selected: 3 \gls{gat} layers with a hidden size of 56 and 4 attention heads each. Note that adding more attention heads could improve the results even further, but it was decided to limit the number to 4 due to memory constraints.

\subsection{Results}
\label{sec:results}

Since the dataset contains dense EDS measurements, the data was modified to simulate the sparse unstructured measurements commonly acquired when fast measurement speed is needed. First, a parameter was chosen that describes a fraction of the spectral data that should be utilised. Based on the parameter, random spectral data points were selected for further processing. This simulates the process where a small set of randomly selected spectral measurements are made. Because of the randomness, the structured nature of the data was effectively removed. Furthermore, this process was done each time data were requested, leading to a new random spectral data selection. This allowed for better utilisation of the provided data and worked as a measure against the overfitting of the model.

\begin{figure}[ht!]
    \centering
    \begin{subfigure}{0.48\columnwidth}
        \includegraphics[width=\linewidth]{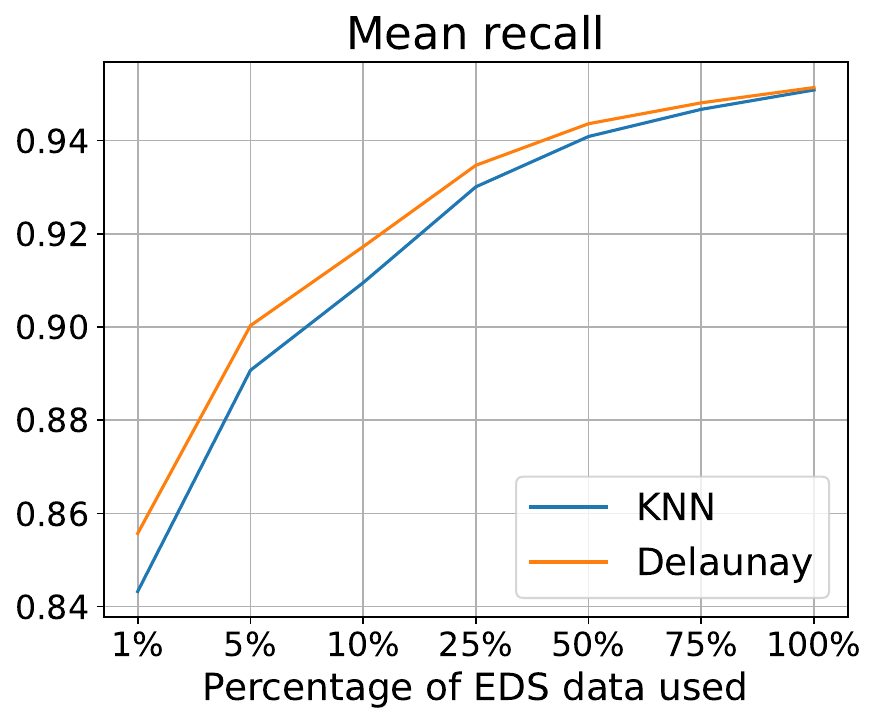}
    \end{subfigure}
    \begin{subfigure}{0.48\columnwidth}
        \includegraphics[width=\linewidth]{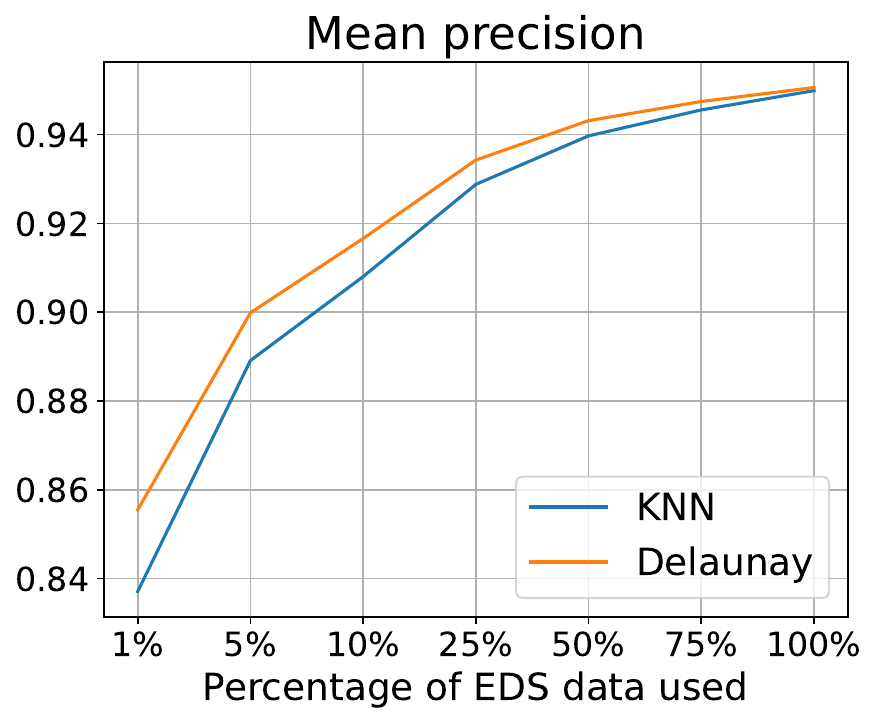}
    \end{subfigure}
    \caption{Evaluation of \gls{kNN} and Delaunay graph construction. Recall and precision of the trained pipeline show that Delaunay triangulation performs better.}
    \label{fig:triangulation_eval}
\end{figure}

The comparison of the two approaches for constructing the EDS part of the graph, Delaunay triangulation (DT) and \gls{kNN}, is shown in Fig. \ref{fig:triangulation_eval}. DT clearly outperformed \gls{kNN} and was selected as the \gls{eds} graph construction method for the rest of the experiments.

\begin{figure}[htp!]
    \centering
    \includegraphics[width=0.9\columnwidth]{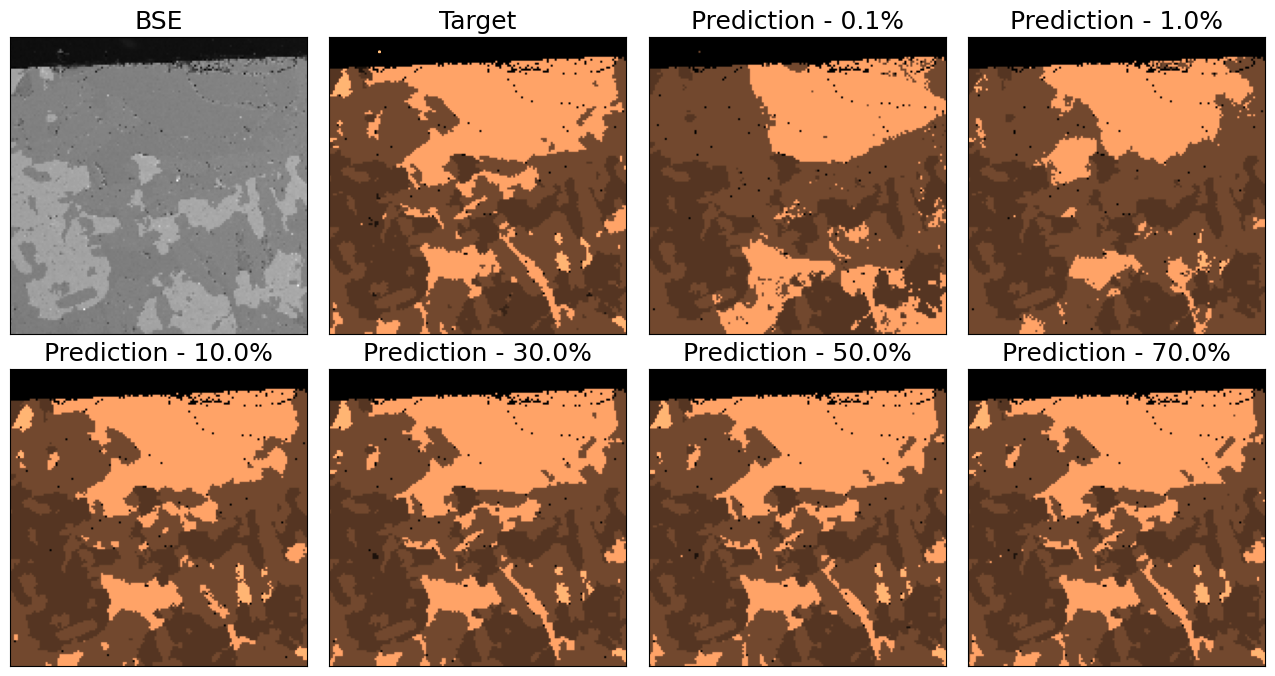}
    \caption{Visualization of outputs: Top left shows \gls{bse} image; the right shows ground truth; others display outputs with varying \gls{eds} data percentages.} 
    \label{fig:ex1_example}
\end{figure}

Example outputs can be seen in Fig.~\ref{fig:ex1_example}.
The model exhibits clear signs of successful data fusion. The figure shows the input \gls{bse} image, as well as the ground truth for the modality pair. The rest are generated segmentations with different percentages of \gls{eds} modality used. It can be seen, that segmentation produced with just 0.1\% of \gls{eds} data is inaccurate, but the results are quickly improving with added data.

\begin{figure}[ht!]
    \centering
    \includegraphics[width=1\columnwidth]{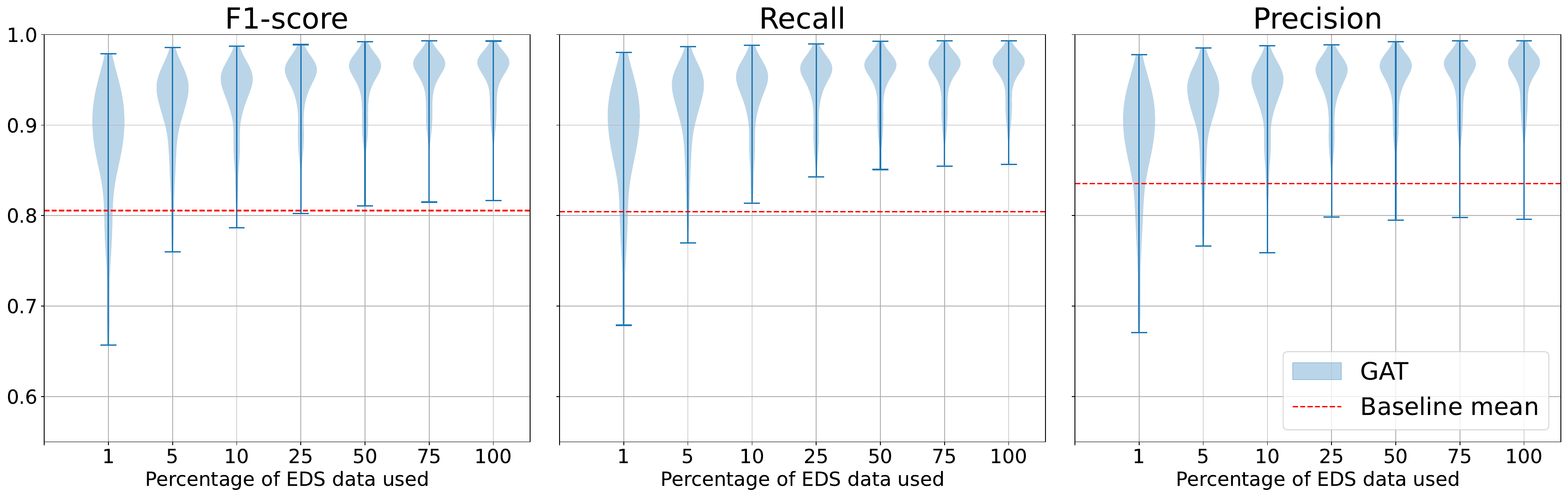}
\caption{Evaluation of the proposed method with different fractions of input \gls{eds} data.}
\label{fig:evaluation_ours}
\end{figure}

\begin{figure}[ht]
    \centering
    {\tiny \hspace{-0.55cm} BSE + EDS sample points \hspace{1.7cm} Target \hspace{2.3cm} Prediction} 
    \includegraphics[width=1\linewidth,trim={0 0.2cm 0 0.72cm},clip]{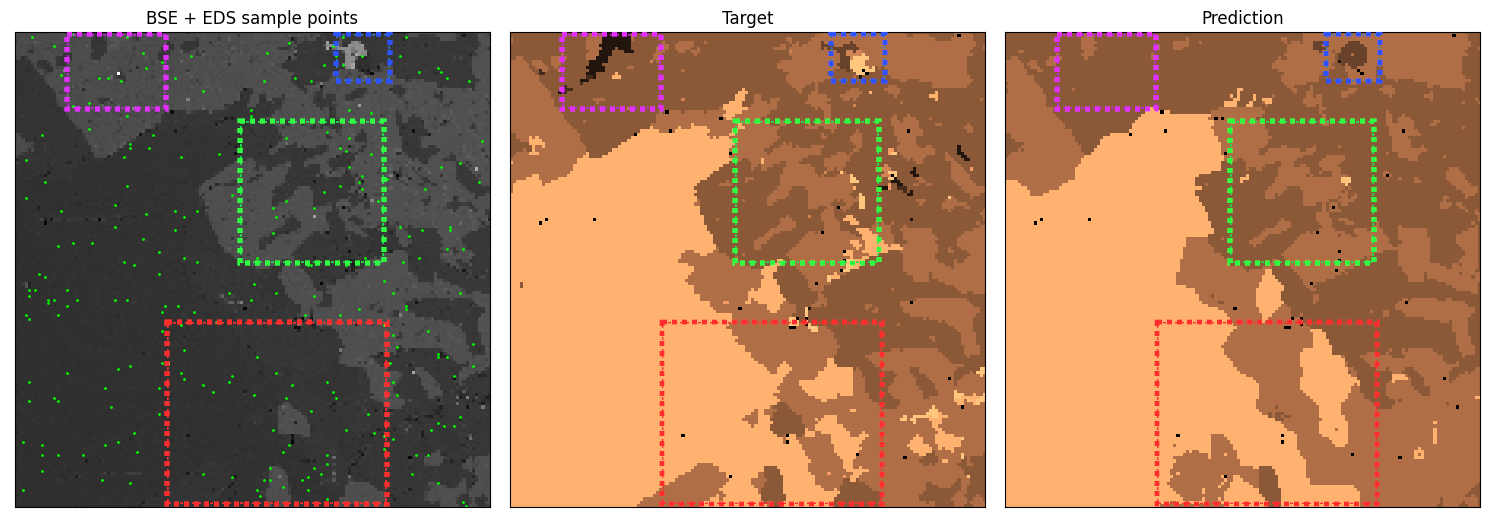}
    \caption{A single prediction using 1\% of \gls{eds} data is shown. Green points indicate \gls{eds} measurement locations. The magenta frame highlights a class missed by the prediction, while the green frame shows an area of high accuracy. The blue frame highlights an area where the distribution of \gls{eds} points caused incorrect prediction, and the red frame marks an area where two similar classes were not separated accurately due to the BSE image's limitations.} 
    \label{fig:highlights}
\end{figure}

The results are shown in Fig.~\ref{fig:evaluation_ours} as well as in Table \ref{tab:comparison}. It can be seen that the model is able to learn from the \gls{eds} data, as the performance of the model increases with the percentage of \gls{eds} data. With only 1\% of \gls{eds} data, the model can achieve promising results in all evaluation metrics. An illustration of such a prediction can be seen in Fig.~\ref{fig:highlights}. With added data, the performance increases in a logarithmic fashion. Note that the $x$-axis is not linear. The proposed method generally outperforms the baseline with 5\% of \gls{eds} data.  Confusion matrices illustrating mineral class predictions of UNet and of the proposed method with 5\% of \gls{eds} data can be seen in the Supplementary material. The matrices show the concrete mineral classes and illustrate the benefits of data fusion.

\begin{table}[ht]
\caption{Segmentation results with various \gls{eds} percentages.}
\centering
\resizebox{0.9\columnwidth}{!}{%
\begin{tabular}{@{}lllllllll@{}}
\toprule
          & UNet & \multicolumn{3}{c}{GDLS} & \multicolumn{3}{c}{Proposed} \\
EDS (\%)  & - & 1\% & 5\% & 10\% & 1\% & 5\% & 10\%  \\ \midrule
Precision & .806 & .743 & .836 & .871 & .879 & .917 & .930  \\
Recall    & .804 & .687 & .797 & .838 & .882 & .920 & .933  \\
F1-score  & .835 & .701 & .811 & .851 & .877 & .917 & .931  \\ \bottomrule
\end{tabular}}
\label{tab:comparison}
\end{table}

\subsubsection*{Comparison with \glsfirst{gdls}}
The proposed method was compared to the \gls{gdls} method for \gls{sem} image segmentation \citep{Juranek_rgbEDS_2022}.
It is important to note that various assumptions had to be made to make the methods comparable. First, as mentioned previously, \gls{gdls} produces class-agnostic segments, whereas the proposed method provides direct segmentation to mineral phases. To alleviate this problem, the output of the \gls{gdls} was sent to the Tescan company, where the segments were transformed to the representation matching the output of the proposed method. This process required a second access to the full \gls{eds} data. This favoured the \gls{gdls} as it had access to a larger amount of \gls{eds} data for segment labelling than the proposed method. Furthermore, \gls{gdls} has two parameters influencing the final segmentation. The parameters were selected on a few samples and used for the whole dataset in such a way which prioritized a higher number of segments. Other parameters could be more suitable, but the selected ones should be enough to provide a good comparison. 
Lastly, the proposed method is not nearly as general as \gls{gdls}, as it was trained on a dataset with a limited number of classes.

\begin{figure}[htp]
    \centering
    \begin{subfigure}{0.43\columnwidth}
        \includegraphics[width=\linewidth]{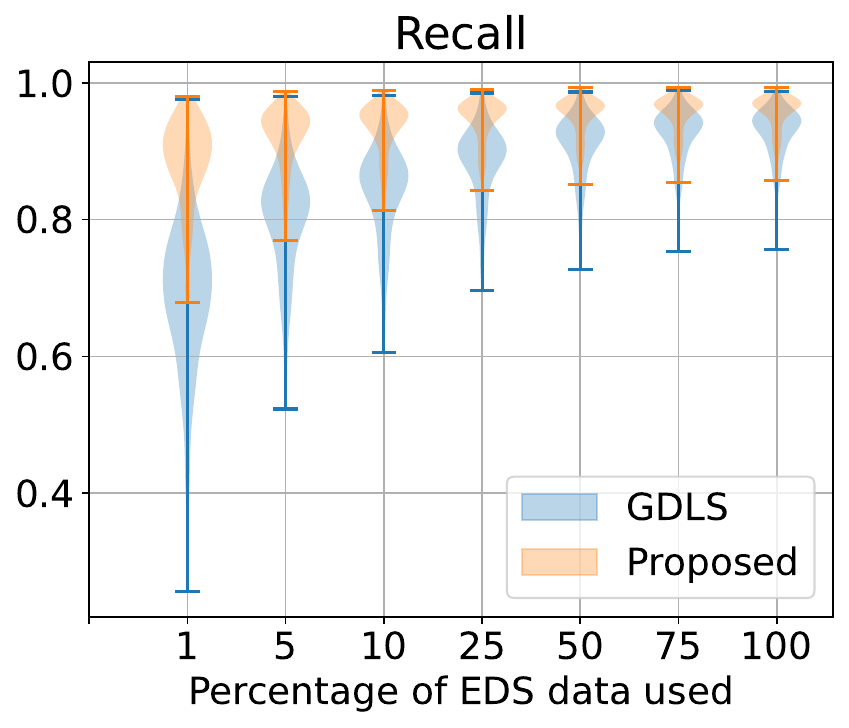}
    \end{subfigure}
    \begin{subfigure}{0.43\columnwidth}
        \includegraphics[width=\linewidth]{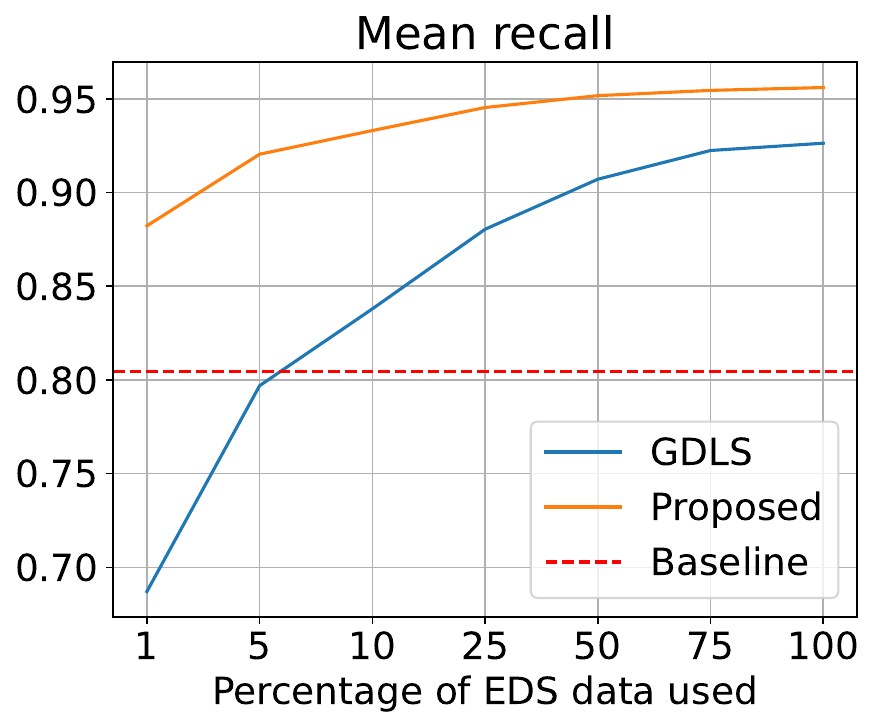}
    \end{subfigure}
    \begin{subfigure}{0.43\columnwidth}
            \includegraphics[width=\linewidth]{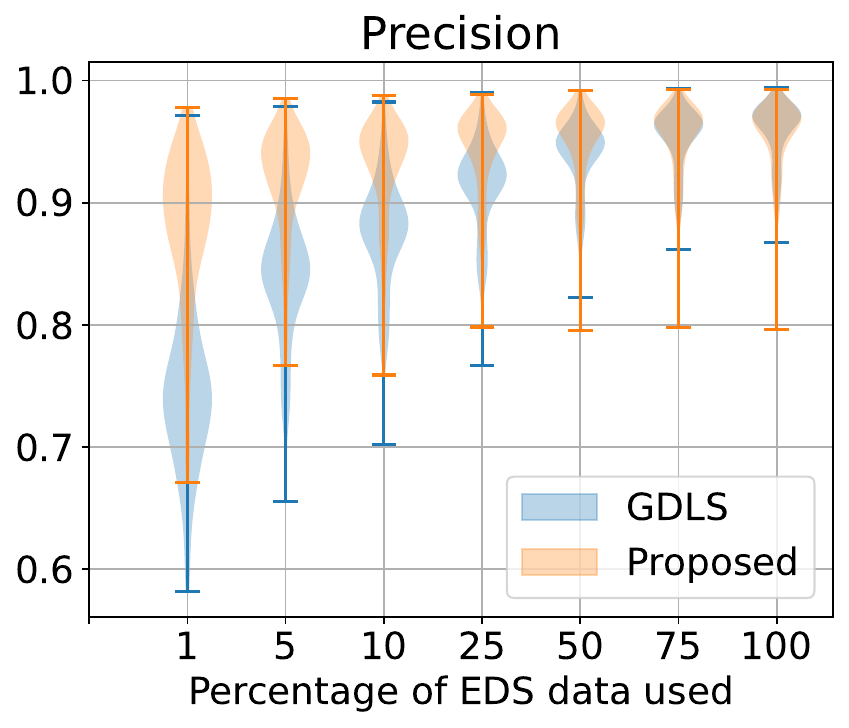}
    \end{subfigure}
    \begin{subfigure}{0.43\columnwidth}
        \includegraphics[width=\linewidth]{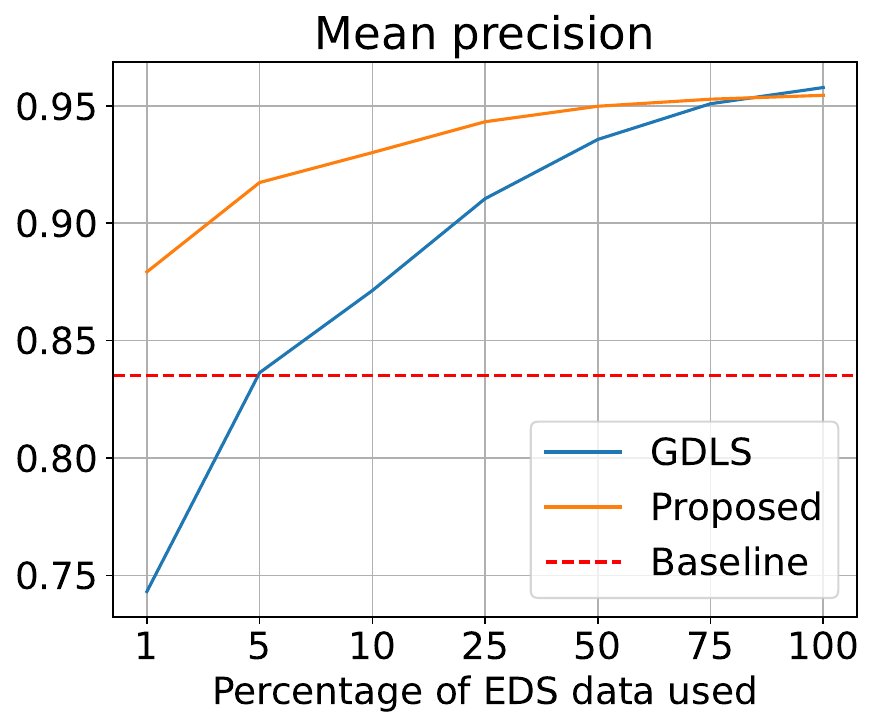}
    \end{subfigure}
    \begin{subfigure}{0.43\columnwidth}
            \includegraphics[width=\linewidth]{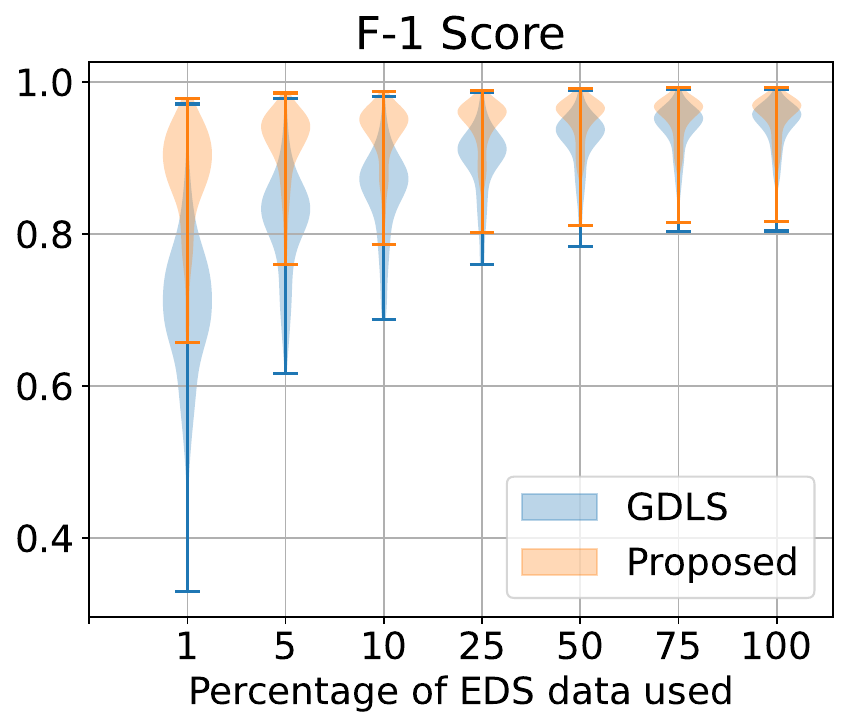}
    \end{subfigure}
    \begin{subfigure}{0.43\columnwidth}
        \includegraphics[width=\linewidth]{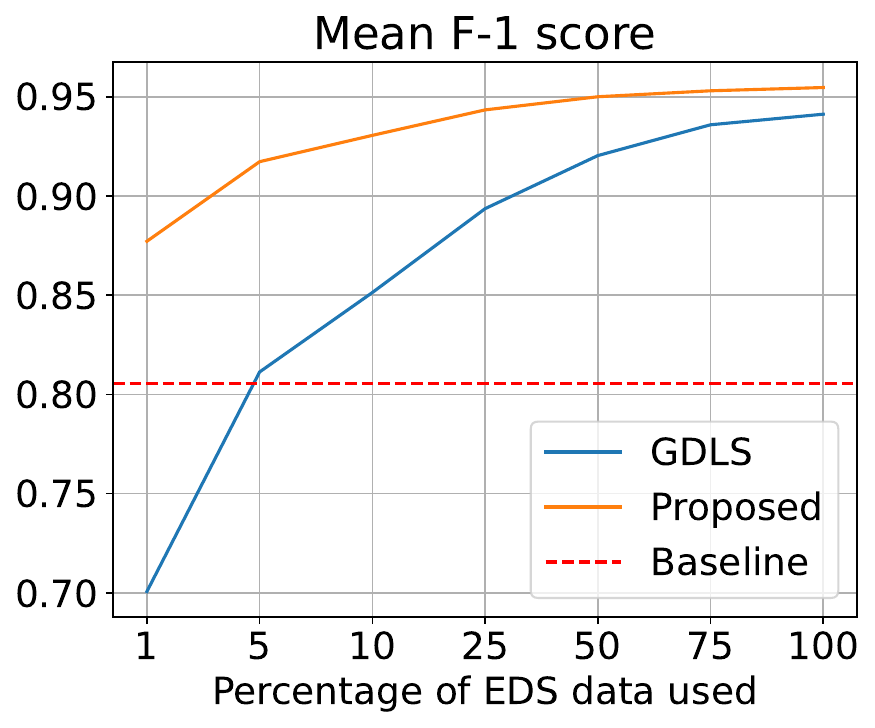}
    \end{subfigure}    
    
    \caption{Evaluation results with comparison of the proposed method to \gls{gdls}.}
    \label{fig:comparison_gdls}
\end{figure}

\begin{figure}[ht]
    \centering
    \includegraphics[width=0.9\linewidth]{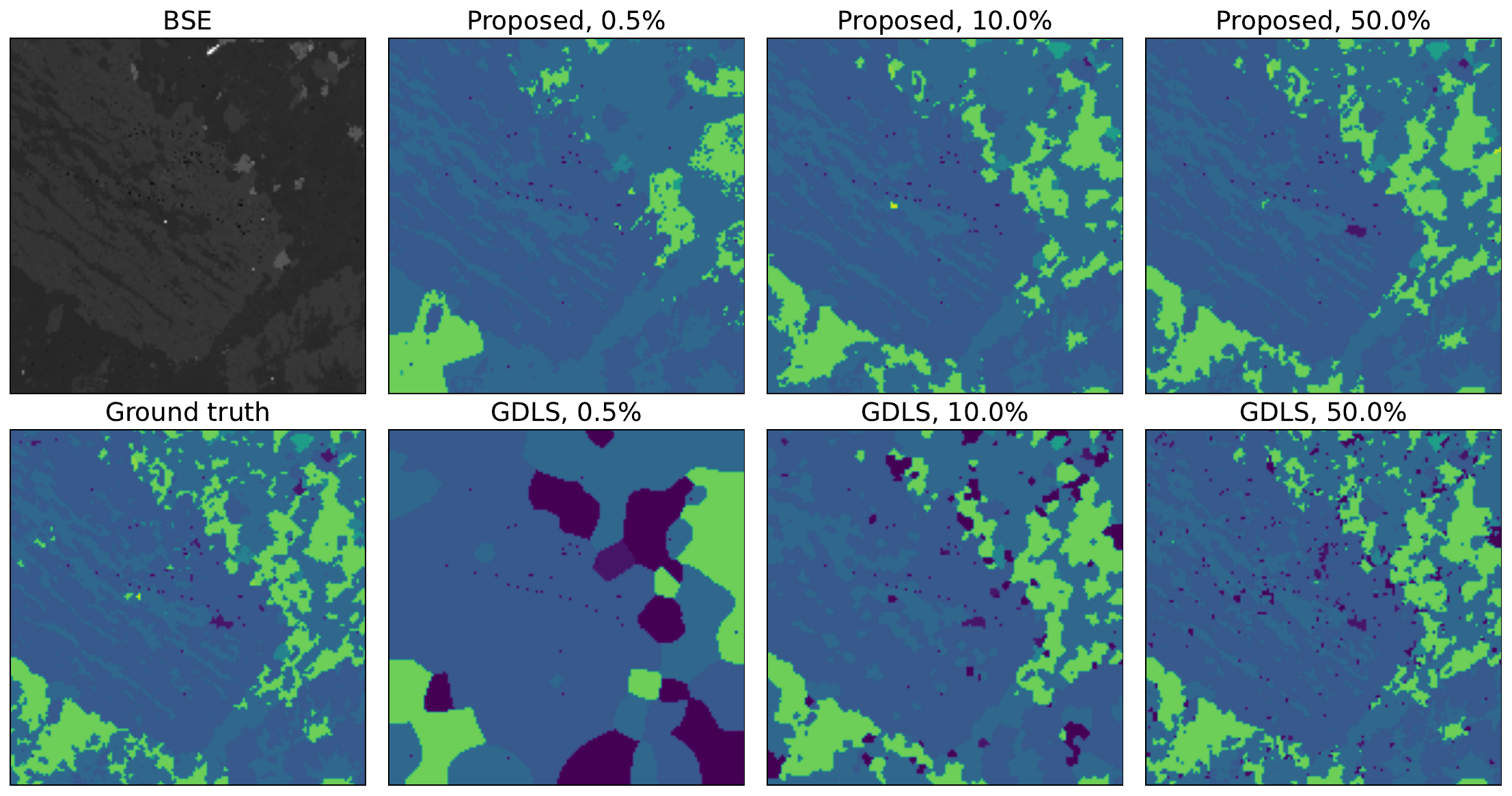}
    \caption{Example outputs of the proposed method and \gls{gdls} for various \gls{eds} data percentages. The leftmost column shows the input \gls{bse} image and the ground truth. The top row shows the output of the proposed method, and the bottom row shows the output of the \gls{gdls} with the same data. } 
    \label{fig:gdls_comparison}
\end{figure}

Evaluation with recall metric shows that the proposed method significantly outperforms the \gls{gdls} in all cases.
However, with precision, the result is not as definite. With a lower number of \gls{eds} data points, the proposed method produces better results; however, with an increasing number of data points, the situation changes. In the highest counts, the \gls{gdls} performance is better than the proposed method. The F1 score shows that the proposed method performs better than the \gls{gdls}. The results can be seen in Fig.~\ref{fig:comparison_gdls}. Example outputs of both methods side by side can be seen in Fig.~\ref{fig:gdls_comparison}. All in all, the experiments show that the method effectively fuses data and is able to outperform the state-of-the-art method.

\section{Conclusion}
\label{sec:conclusion}

In this paper, we presented a simple yet versatile graph-based method for data fusion and multimodal image segmentation. The method constructs a joint graph representation of the input modalities and processes the graph using \gls{gat}. The main advantage of this method is its flexibility; it does not require the modalities to have a predefined structure, such as a grid-like arrangement. Instead, it can be applied to a wide variety of modalities, including images, point clouds, and point-wise measurements. We demonstrate the effectiveness of our method on \gls{sem} mineral segmentation using \gls{bse} images and sparse \gls{eds} measurements. Our results show that the proposed method outperforms competing methods. We consider the most valuable scientific contribution of the paper to be experimental proof of the feasibility of the proposed multimodal fusion of heterogeneous input data. The method, including the graph construction and processing steps, is general and can be applied to other applications involving image data and non-grid-like measurements. 

Future work includes the use of heterogeneous graphs or expanding to class-agnostic segmentation.
\section*{Acknowledgments}
This work was supported by the Finnish Ministry of Education and Culture’s Pilot for Doctoral Programmes (Pilot project Mathematics of Sensing, Imaging and Modelling). We would like to express our gratitude to Tescan Group company for their generous provision of data. Additionally, we extend our heartfelt thanks to David Motl for guidance and expertise.
\subsection*{CRediT authorship contribution statement}
\noindent
\textbf{Samuel Repka} - Software, Validation, Writing - Original draft,
\textbf{Bořek Reich} \& \textbf{Fedor Zolotarev} - Supervision, Writing - Review \& Editing,
\textbf{Tuomas Eerola} \& \textbf{Pavel Zemčík} - Conceptualization, Supervision, Writing - Review \& Editing

\bibliographystyle{model5-names}
\bibliography{bibliography}



\end{document}


\noindent\Large{Supplementary material for:\\Mineral segmentation using electron microscope images and spectral sampling through
multimodal graph neural networks}

\vspace{10pt}

\noindent\normalsize{Samuel Repka, Bořek Reich, Fedor Zolotarev, Tuomas Eerola, Pavel Zemčík}

\section*{Confusion matrices}
\begin{figure}[ht!]
    \centering
        \includegraphics[width=0.9\linewidth]{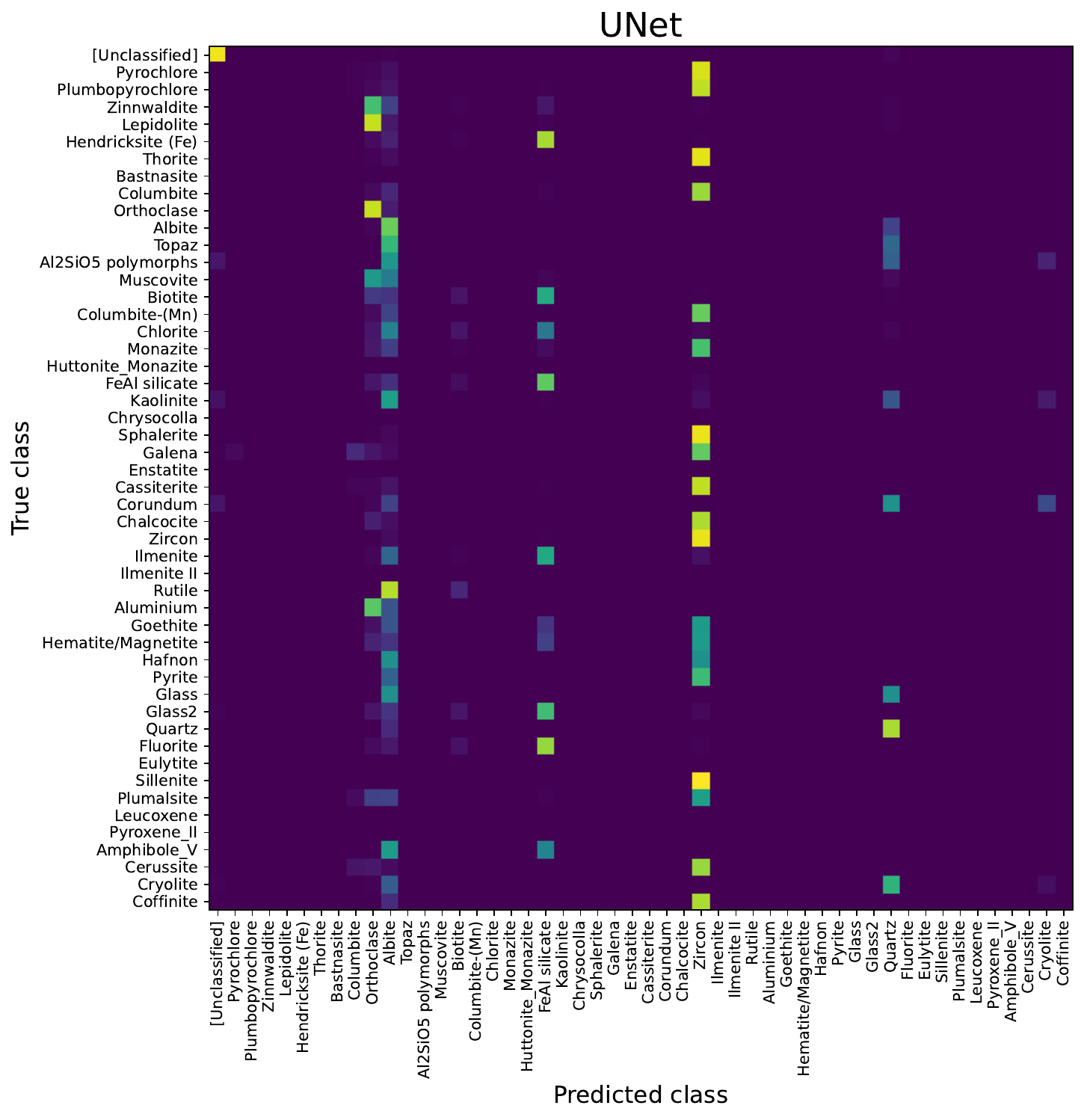}
    \caption{Confusion matrix for the UNet baseline.}
    \label{fig:confmat_unet}
\end{figure}

\begin{figure}[ht!]
    \centering
        \includegraphics[width=0.9\linewidth]{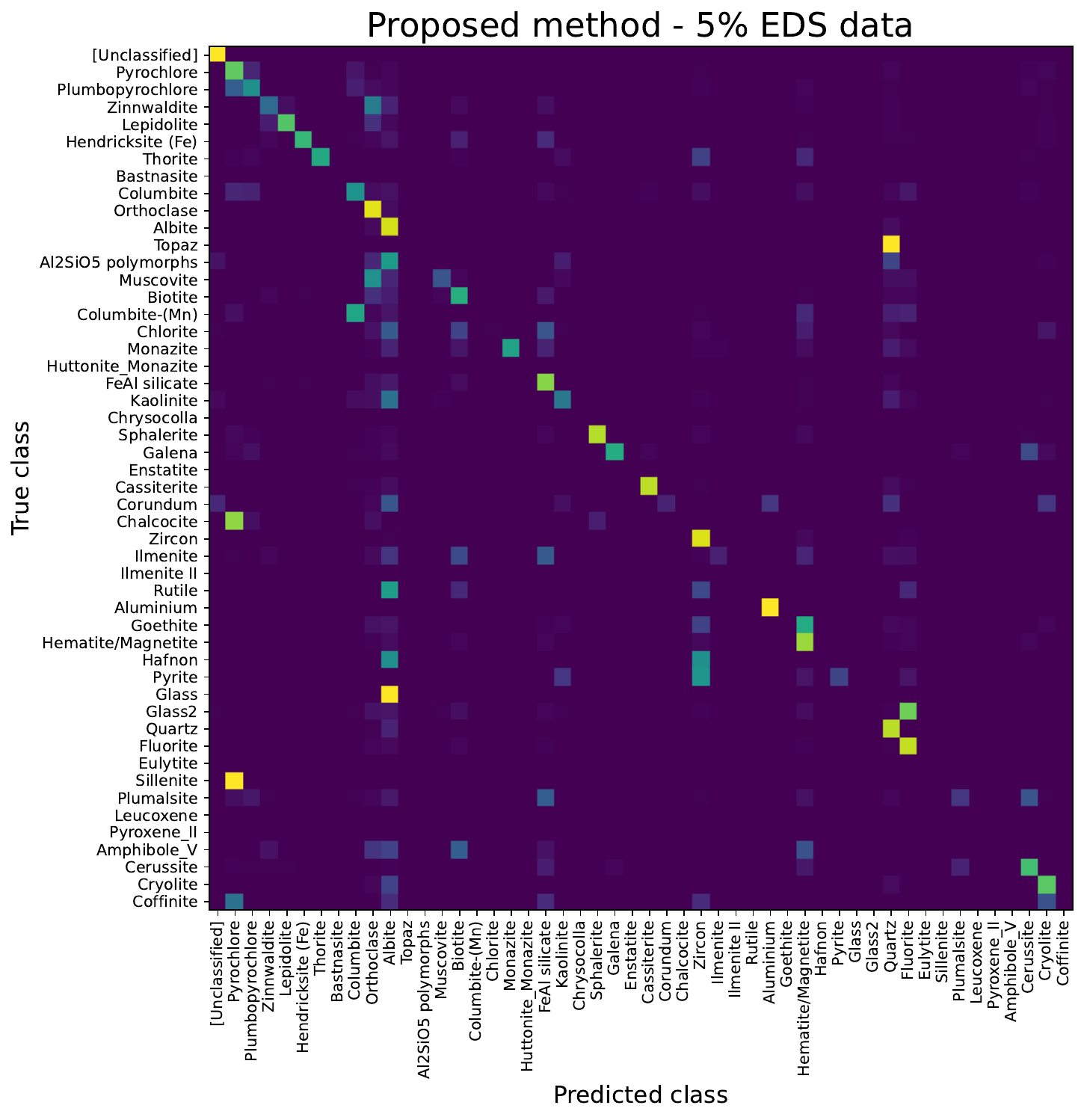}
    \caption{Confusion matrix for the propoed method.}
    \label{fig:confmat_gnn}
\end{figure}

Confusion matrices for the UNet baseline and proposed method are shown in Figures~\ref{fig:confmat_unet} and~\ref{fig:confmat_gnn} . While UNet provides reasonably good numerical results, the information available in the data is simply insufficient for reliable predictions. The proposed method predictions are much better, as it is able to utilise much more feature-rich \gls{eds} data.

